\documentclass[draftclsnofoot,11pt,journal,onecolumn]{IEEEtran}
\usepackage[top=1in,bottom=1in,left=1in,right=1in]{geometry}
\usepackage[utf8]{inputenc}
\usepackage{cite}
\usepackage{mathtools}
\usepackage{amssymb}
\usepackage{amsthm}
\usepackage{balance}
\usepackage{graphicx}
\usepackage{epstopdf}
\usepackage{algpseudocode}
\usepackage[usenames]{xcolor}
\usepackage{algorithmicx,algorithm}
\usepackage{fancybox}
\usepackage{tablefootnote}
\usepackage{mdframed}
\usepackage{wrapfig}

\newcommand{\E}         {\mathbb{E}}

\newcommand{\bg}        {\mathbf{g}}

\newcommand{\bh}        {\mathbf{h}}
\newcommand{\bH}        {\mathbf{H}}
\newcommand{\cH}        {\mathcal{H}}

\newcommand{\cN}        {\mathcal{N}}

\newcommand{\cO}        {\mathcal{O}}

\newcommand{\cP}        {\mathcal{P}}

\newcommand{\R}         {\mathbb{R}}

\newcommand{\bw}        {\mathbf{w}}

\newcommand{\wtbw}      {\widetilde{\bw}}

\newcommand{\whbw}      {\widehat{\bw}}

\newcommand{\by}        {\mathbf{y}}

\newcommand{\bz}        {\mathbf{z}}
\newcommand{\bZ}        {\mathbf{Z}}

\newcommand{\bEta}      {\boldsymbol{\eta}}

\DeclareMathOperator*{\argmin}  {arg\,min}

\DeclarePairedDelimiter{\ceil}{\lceil}{\rceil}

\mdfdefinestyle{MyFrame}{%
    linecolor=black,
    linewidth=2pt,
    leftmargin=0pt,
    rightmargin=0pt,
    backgroundcolor=gray!15,
    innertopmargin=4pt,
    innerbottommargin=4pt,
    innerrightmargin=4pt,
    innerleftmargin=4pt,
}

\title{\huge Adversary-resilient Distributed and Decentralized Statistical Inference and Machine Learning\\ \Large [An Overview of Recent Advances Under the Byzantine Threat Model]}
\author{Zhixiong~Yang, Arpita~Gang, and Waheed~U.~Bajwa}

\begin{document}
\maketitle

Statistical inference and machine learning algorithms have traditionally been developed for data available at a single location. Unlike this \emph{centralized setting}, modern datasets are increasingly being distributed across multiple physical entities (sensors, devices, machines, data centers, etc.) for a multitude of reasons that range from storage, memory, and computational constraints to privacy concerns and engineering needs. This has necessitated the development of inference and learning algorithms capable of operating on non-collocated data. We divide such algorithms into two broad categories in this article, namely, \emph{distributed algorithms} and \emph{decentralized algorithms}.

\textbf{Distributed Algorithms:} Distributed algorithms correspond to setups in which the data-bearing entities (henceforth referred to as ``nodes'') require some form of coordination through a specially designated entity in the system in order to generate the final result. Depending on the application, this special entity is referred to as master node, central server, parameter server, fusion center, etc., in the literature. While distributed setups can take a number of forms, this exposition mostly revolves around the so-called \emph{master--worker distributed architecture} in which data-bearing nodes only communicate with a single master node that is tasked with generating the final result. Among other applications, such distributed architectures arise in the context of parallel computing, where the focus is computational speedups and/or overcoming memory/storage bottlenecks, and federated systems, where ``raw'' data collected by individual nodes cannot be shared with the master node due to either communication constraints (e.g., sensor networks) or privacy concerns (e.g., smartphone data).

\textbf{Decentralized Algorithms:} Decentralized algorithms correspond to setups that lack central servers; instead, data-bearing nodes in a decentralized system are \emph{collectively} tasked with generating the final result. In particular, individual nodes in a decentralized setup typically communicate among themselves over a network (often ad hoc) to reach a common solution (i.e., achieve \emph{consensus}) at all nodes. Decentralized setups arise either out of the need to eliminate single points of failure in distributed setups or due to practical engineering constraints, as in the internet of things and autonomous systems.

\begin{mdframed}[style=MyFrame]
\emph{Is it distributed or is it decentralized?} Inference and learning from non-collocated data have been studied for decades in computer science, control, signal processing, and statistics. Both among and within these disciplines, however, there is no consensus on use of the terms ``distributed'' and ``decentralized.'' Though many works share the definitions provided in here, there are numerous authors who use these two terms interchangeably, while there are some other authors who reverse these definitions.
\end{mdframed}

\begin{figure}
    \centering
    \includegraphics[width=\textwidth]{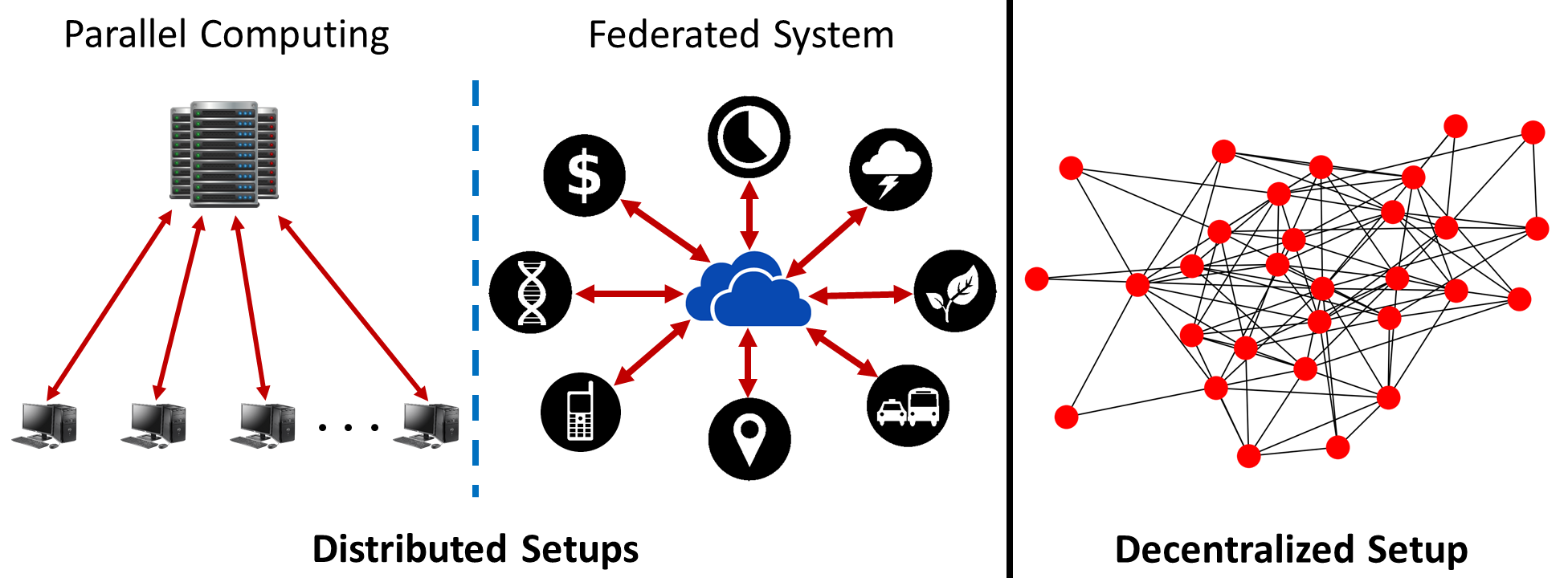}
    \caption{\linespread{1}\selectfont\footnotesize Inference and machine learning algorithms involving non-collocated data can be broadly divided into the categories of ($i$) distributed algorithms and ($ii$) decentralized algorithms. The former category includes master--worker distributed setups such as parallel computing and federated systems in which a single entity is tasked with generating the final result. The latter category deals with decentralized setups in which entities cooperate among themselves to produce the final result.}
    \label{fig:distributed.vs.decentralized}
\end{figure}

The last few decades have witnessed the emergence of a plethora of applications that necessitate advances in inference and learning in distributed and decentralized setups. Indeed, distributed and decentralized statistical inference methods are increasingly being relied upon in urban traffic monitoring, environmental sensing, management of smart grids, distributed spectrum sensing, and homeland security, among other applications. Similarly, distributed and decentralized machine learning algorithms are increasingly being utilized in the context of networks of self-driving cars, control of robot swarms, pattern recognition in large-scale datasets, and federated learning systems for healthcare data. Collectively, these applications have resulted in the development of a huge body of work devoted to understanding the algorithmic and theoretical underpinnings of distributed and decentralized inference and learning. But much of this work assumes a non-adversarial setting in which individual nodes---apart from occasional statistical failures---operate as intended within the algorithmic framework.

In recent years, however, cybersecurity threats from malicious non-state actors and rogue entities---and the potentially disastrous consequences of these threats for the aforementioned applications---have forced practitioners and researchers to rethink the robustness of distributed and decentralized algorithms against adversarial attacks. As a result, we now have an abundance of algorithmic approaches that guarantee robustness of distributed and/or decentralized inference and learning under different adversarial threat models; see, e.g., recent survey articles \cite{vempaty.varshney2013article,chen.moura2018article,zhang.poor2018article}. Driven in part by the world's growing appetite for data-driven decision making, however, securing of distributed/decentralized frameworks for inference and learning against adversarial threats remains a rapidly evolving research area. In this article, we provide an overview of some of the most recent developments in this area under the threat model of \emph{Byzantine attacks}. This threat model---which subsumes the case of malfunctioning nodes, referred to as \emph{Byzantine faults/failures}---is one of the hardest to safeguard against since it allows for undetected takeover of nodes by the adversary. In particular, nodes affected under this threat model---termed \emph{Byzantine nodes}---are assumed to have the potential to arbitrarily bias the outputs of the underlying algorithms by colluding among themselves, injecting false data and information into the distributed/decentralized system, etc. This is in stark contrast to the relatively simpler threat model of \emph{crash faults}, in which individual nodes in the distributed/decentralized system continue to operate as intended till a crash fault occurs, at which point the faulty node ceases to interact with the system (rather then potentially injecting false information into the system). We refer the reader to~\cite{Su2015fault}, and the references within, for further discussion on both the generality and the hardness of the Byzantine threat model in relation to the crash-fault model.

\begin{mdframed}[style=MyFrame]
\begin{wrapfigure}{r}{5.5cm}
    \vspace{-\baselineskip}
	\includegraphics[width=5.5cm]{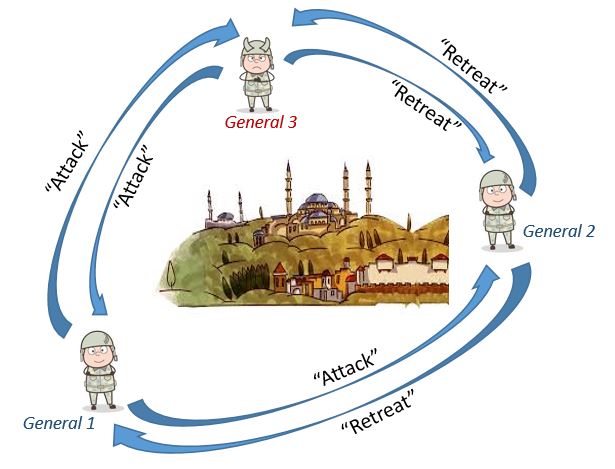}
	\vspace{-1.75\baselineskip}
	\caption{\linespread{1}\selectfont\footnotesize A Byzantine army led by three generals, one of whom (General~3) is a traitor, surrounding an enemy city. The loyal generals are trying to reach a consensus on the plan of action against the enemy, while the traitor is trying to mislead them.}
	\vspace{-\baselineskip}
	\label{fig:bgp}
\end{wrapfigure}
\emph{Origination of the Byzantine threat model:} The threat model of Byzantine attacks/faults/failures in its most general form was introduced and analyzed in \cite{Lamport1982Byzantine} within the context of reliability of computer systems with potentially malfunctioning components. The overarching problem in \cite{Lamport1982Byzantine} was abstracted as the \emph{Byzantine Generals Problem}, in which several generals of the Byzantine army---some of whom are likely to be traitors---need to agree on attacking or retreating from an enemy city through exchange of messages via a messenger; see, e.g., Figure~\ref{fig:bgp}. The authors reported two key results in \cite{Lamport1982Byzantine} for this problem. First, they established the impossibility of safeguarding against Byzantine nodes (traitor generals) when the number of uncompromised nodes (loyal generals) is not more than two-thirds of the total number of nodes. The threat from General~3 in Figure~\ref{fig:bgp}, therefore, cannot be neutralized by any algorithm since the number of loyal generals is only two in this scenario. Second, the authors proposed two algorithms for \emph{complete} and \emph{regular} graphs that provably counter the actions of Byzantine nodes in the case of binary (e.g., attack/retreat) decisions as long as the number of uncompromised nodes exceeds two-thirds of the total number of nodes.%
\end{mdframed}

The Byzantine threat model, since its introduction in 1982, has been extensively studied in the context of statistical inference from non-collocated data~\cite{vempaty.varshney2013article,chen.moura2018article}. However, its potential impact on more general decentralized consensus and distributed/decentralized machine learning problems has only been recently studied. It is against this backdrop that this article summarizes recent developments in Byzantine-resilient processing of non-collocated data, with a majority of the discussion focused on machine learning problems in distributed and decentralized settings.

\section{Adversary-resilient Distributed Processing of Data}
In the classic master--worker setting of distributed systems tasked with processing of non-collocated data, there is one central server---referred to as master node, parameter server, fusion center, etc.---that coordinates with $M$ data-bearing nodes (sensors, smartphones, worker nodes, etc.) for final decision making. While all nodes in this setting communicate with the server, they usually cannot communicate with each other (see ``Distributed Setups'' in Figure~\ref{fig:distributed.vs.decentralized}). The Byzantine threat model in this setting assumes at most $b$ nodes in the system have been compromised. This parameter $b$, which typically corresponds to a crude upper bound on the exact number of Byzantine nodes, often plays an important role in both the analysis and the performance of Byzantine-resilient algorithms. We summarize some of these algorithms and their theoretical guarantees in the following for statistical inference and machine learning problems.

\subsection{Distributed Statistical Inference}
Statistical inference leverages data samples in order to draw conclusions about the underlying probability distribution(s) generating the data. While statistical inference can take many forms, we limit our discussion in this article to Byzantine-resilient distributed detection and distributed estimation.

\textbf{Distributed Detection:} Distributed detection under both the \emph{Neyman--Pearson} (NP) and the \emph{Bayesian} frameworks has a rich history. Given two hypotheses $\cH_0$ and $\cH_1$, a typical distributed detection algorithm first involves each node taking a local decision in favor of either $\cH_0$ or $\cH_1$ based on its own data samples. The nodes then send their decisions to the central server, which applies a fusion rule to the local decisions in order to reach the final (global) decision. Unfortunately, distributed detection algorithms designed without consideration of potential Byzantine failures break down in the presence of Byzantine nodes.

Despite the brittleness of traditional distributed detection techniques, investigation of Byzantine-resilient distributed detection only took off in the last decade. The survey article~\cite{vempaty.varshney2013article} provides an overview of many of the resulting methods and asymptotically characterizes the respective \emph{critical fraction} $\alpha^*$ of Byzantine nodes, defined as $\alpha^* := \tfrac{b}{M}$, at (or beyond) which the central server cannot do better than random guessing for its final decision. An important insight from the earliest works on Byzantine-resilient distributed detection is that (asymptotically) $\alpha^*$ can be $\tfrac{1}{2}$ or higher~\cite{vempaty.varshney2013article}; in contrast, recall that $\alpha^* = \tfrac{1}{3}$ for the original Byzantine Generals Problem~\cite{Lamport1982Byzantine}.

Since the appearance of \cite{vempaty.varshney2013article} in 2013, a few other works on Byzantine-resilient distributed detection have appeared in the literature. It is shown in~\cite{nadendla.varshney2014} that distributed detection can be more resilient to Byzantine failures in the case of a general $Q$-ary hypothesis testing problem, with the critical fraction given by $\alpha^* = \tfrac{Q-1}{Q}$. Byzantine-resilient distributed binary hypothesis testing is investigated for the first time under the Bayesian framework in~\cite{kailkhura.varshney2015trans}, with the critical fraction also given by $\alpha^* = \tfrac{1}{2}$ in this case. Finally, it is established in~\cite{hashlamoun.varshney2018} that---under certain conditions---it is possible to have $\alpha^* = 1$ in the case of Bayesian distributed binary detection as long as each node is allowed to replicate its message to the central server through one other node in the system. Strictly speaking, however, this coordination among pairs of nodes leads to a distributed architecture that differs from the distributed master--worker (star topology) architecture of prior works. We conclude by noting that a tree topology in which the central server sits at the root of the tree (Depth 0) and nodes in the distributed system route their messages to the server through their parent nodes is another distributed architecture that does not fall under the distributed master--worker setup. Byzantine resilience of such architectures in distributed detection tasks is investigated in~\cite{kailkhura.varshney2015}, with the critical fraction of Byzantine nodes defined in terms of the number of Byzantine nodes $b_1$ and the total number of nodes $M_1$ at Depth 1 of the tree. We also refer the reader to Table~\ref{table: distributed_detection} for a summary of all the results that have been discussed in this article in relation to Byzantine-resilient distributed detection.

\begin{table}
\caption{Summary of recent results concerning Byzantine-resilient distributed detection} \label{table: distributed_detection}
\vspace{-\baselineskip}
\begin{center}
\begin{tabular}{|l|l|l|l|l|}
\hline
\textbf{Reference} & \textbf{Hypothesis Test}  & \textbf{Detection Setup} & \textbf{Critical Fraction $\alpha^*$} & \textbf{Distributed Architecture}\\
\hline\hline
Nadendla et al.~\cite{nadendla.varshney2014} & $Q$-ary & Neyman--Pearson & $\alpha^* = \frac{Q-1}{Q}$ & Star topology\\
\hline
Kailkhura et al.~\cite{kailkhura.varshney2015trans} & Binary & Bayesian & $\alpha^* = 0.5$ & Star topology\\
\hline
Hashlamoun et al.~\cite{hashlamoun.varshney2018} & Binary & Bayesian & Can have $\alpha^* = 1$ & Quasi-star topology\\
\hline
Kailkhura et al.~\cite{kailkhura.varshney2015} & Binary & Neyman--Pearson & $b_1 = \ceil{\frac{M_1}{2}}$ & Tree topology\\
\hline
\end{tabular}
\end{center}
\vspace{-1.25\baselineskip}
\end{table}

\textbf{Distributed Estimation:} Byzantine-resilient distributed estimation has received significant attention lately in the context of state estimation in cyberphysical systems. Much of the developments in this regard have been limited to linear models, with a typical observation model at any given time at node $j$ expressed as $y_j = \bh_j^T \bw + \eta_j, \ j = 1,\dots,M$, where $\bh_j \in \R^d$ denotes a known vector, $\bw \in \R^d$ is the unknown state vector that needs to be estimated at the central server, and $\eta_j$ denotes the observation noise at node $j$. Traditionally, distributed estimation has involved the nodes transmitting $y_j$'s (or their quantized versions) to the central server and the server estimating $\bw$ using some variant of the following least-squares formulation~\cite{HauptBajwaEtAl.ISPM08} (or a maximum likelihood one in the case of quantized transmissions):
\begin{align}
    \label{eqn:dist.state.est}
    \whbw = \argmin_{\wtbw} \big\|\by - \bH\wtbw\big\|_2,
\end{align}
where the matrix $\bH \in \R^{M \times d}$ has $\bh_j$'s as its rows. The quality of such solutions are assessed in terms of gap of their \emph{mean squared error} (MSE), $\E[\|\whbw - \bw\|_2^2]$, from either the \emph{minimum mean squared error} (MMSE) or the \emph{Cramer--Rao lower bound} (CRLB) for unbiased estimators.

In the presence of Byzantine nodes, in which case the $y_j$'s corresponding to the $b$ Byzantine nodes can be arbitrarily different from $\bh_j^T \bw$, solutions of distributed estimation techniques based on \eqref{eqn:dist.state.est} can be pushed far from the optimal (in terms of MMSE/CRLB); see, e.g., Figure~\ref{fig:dist_est}. In this setting, one can again characterize the critical fraction $\alpha^*$ of Byzantine nodes at (or beyond) which the server can do no better than relying on prior knowledge about $\bw$. It is, for instance, established in~\cite{nadendla.varshney2014} that---under certain assumptions---it is possible to have $\alpha^* = \tfrac{Q-1}{Q}$ when nodes utilize $Q$-ary quantization in order to transmit their data to the server. The survey articles~\cite{vempaty.varshney2013article, chen.moura2018article, zhang.poor2018article} provide additional discussion of works on Byzantine-resilient distributed estimation, many of which are based on the idea of either detection of Byzantine attacks and/or identification of individual Byzantine nodes as part of their mitigation strategy. Finally, while much of the focus in Byzantine-resilient distributed estimation has been on linear models, there are works like~\cite{vempaty.agarwal.varshney2013} that focus on nonlinear models. Since there is a significant overlap between statistical estimation and machine learning problems, we do not indulge in discussion of such works; instead, we move towards our review of recent results on Byzantine-resilient distributed machine learning.

\begin{mdframed}[style=MyFrame]
\begin{wrapfigure}{r}{5.5cm}
    \vspace{-\baselineskip}
	\includegraphics[width=5.5cm]{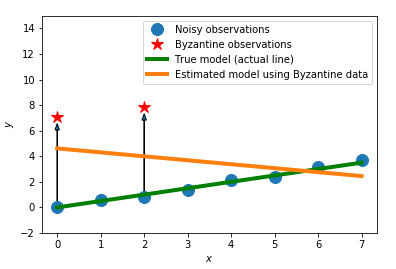}
	\vspace{-1.75\baselineskip}
	\caption{\linespread{1}\selectfont\footnotesize The impact of Byzantine failures on traditional least-squares estimation.}
	\label{fig:dist_est}
	\vspace{-\baselineskip}
\end{wrapfigure}
\emph{Distributed estimation and Byzantine failures:} We illustrate the impact of Byzantine nodes on distributed estimation through a simple example of two-dimensional parameter estimation, in which the linear observation model corresponds to a line in $(x,y)$ plane, using $M = 8$ nodes. The observations at each node in this example correspond to $y_j = \bh_j^T \bw + \eta_j$ with $\bh_j^T := [x_j \ 1]$ and the two-dimensional vector $\bw$ describing the slope and $y$-intercept of the line. It can be seen from Figure~\ref{fig:dist_est} that Byzantine failures of just two nodes returns a least-squares solution that results in a model $\bh_j^T \whbw$ that significantly differs from the true model $\bh_j^T \bw$.
\end{mdframed}

\subsection{Distributed Machine Learning}
A typical challenge in machine learning is to statistically minimize a function $f$, referred to as \emph{loss function} or \emph{risk function}, with respect to a candidate \emph{model} $\bw \in \R^d$ that describes the data, i.e.,
\begin{align}\label{eqn: ML problem}
    \min\limits_{\bw} \E_{\bz \sim \cP}\big[f(\bw,\bz)\big],
\end{align}
where $\bz$ denotes data living in some Hilbert space that is drawn from an \emph{unknown} probability distribution $\cP$. To solve \eqref{eqn: ML problem} without knowledge of $\cP$, distributed machine learning focuses on minimizing an empirical variant of $\E_{\bz \sim \cP}\big[f(\bw,\bz)\big]$ using samples of $\bz$---termed \emph{training data}---distributed across different nodes. The resulting objective, aptly termed distributed \emph{empirical risk minimization} (ERM), can be expressed as $\min\limits_{\bw} \frac{1}{M}\sum\limits_{j=1}^M f(\bw, \bZ_j)$, with $\bZ_j$ denoting the local training data at node $j$ that comprises multiple samples drawn from $\cP$ and $\bw$ referred to as the \emph{global} optimization variable. The model/variable $\bw$ in distributed machine learning is stored at the central server, which iteratively updates it based on messages received from individual nodes and subsequently sends the updated $\bw$ to all nodes. Each node, in turns, performs some computation according to its local data and the received $\bw$, and sends a message back to the server.

\textbf{Distributed Stochastic Gradient Descent:} Similar to distributed statistical inference, Byzantine failures can lead to breakdowns of distributed machine learning methods. Motivated in part by the widespread adoption of deep neural networks and variance-reduction techniques like mini batching by the practitioners, we mainly focus here on robustification of synchronous distributed \emph{stochastic gradient descent} (SGD) against Byzantine attacks. A typical distributed SGD algorithm proceeds iteratively. In each iteration, nodes compute the gradient of the loss function on their local data with respect to the current model and send their gradients to the server. The server, in turn, takes the average of all the gradients and updates the global variable according to the averaged gradient. In a faultless environment, distributed SGD---with proper choices of step size and batch size---has a linear convergence rate for strongly convex functions. However, a single Byzantine node can force the algorithm to converge to any model using a simple strategy. Suppose, for instance, the summation of gradients of faultless nodes is $\bg$ and the Byzantine node wishes the server to operate on an alternate gradient $\bg'$. The Byzantine node can accomplish this by sending $M\bg' - \bg$ as its gradient to the server, thereby controlling the update step at the server. Of course, as simple as this strategy is, it is not an optimal one for the Byzantine node; indeed, a large $M$ will result in a large gradient, which is likely to make the malicious message detectable. We refer the reader to~\cite{xie2019fall,fang2019local,baruch2019little} for more sophisticated strategies that can be employed by Byzantine nodes.

Several algorithms have been put forth recently to safeguard distributed SGD against Byzantine failures. The central idea in all these approaches that imparts Byzantine resilience to distributed SGD involves the use of a \emph{screening} procedure at the server while it aggregates the local gradients. Using an appropriate screening rule, the screened aggregated gradient can be shown to be close to the true average gradient, thereby enabling the server to approximately solve the distributed ERM problem. Algorithm~\ref{alg: SGD} provides a general framework for Byzantine-resilient distributed SGD that is based on this idea of screening. While some algorithms alter this framework a little bit, the general idea remains the same. Note that if one were to remove the screening procedure in Step~\ref{alg_step: screening} of the algorithm, it becomes vanilla distributed SGD.

\begin{algorithm}[h]
\caption{A General Framework for Byzantine-resilient Distributed SGD} \label{alg: SGD}
\begin{algorithmic}[1]
\For{$t=1,2,\dots$}
\State \textbf{Server}:
\State \quad Send the global optimization variable to all nodes \label{alg_step: server send}
\State \textbf{Node}:
\State \quad Receive the global optimization variable from the server \label{alg_step: node receive}
\State \quad Calculate gradient with respect to the local training set
\State \quad Send the local gradient to the server \label{alg_step: node send}
\State \textbf{Server}:
\State \quad Receive local gradients from all nodes \label{alg_step: server receive}
\State \quad Screen the received gradients for Byzantine resilience and aggregate them \label{alg_step: screening}
\State \quad Update the global variable by taking a gradient step using the aggregated gradient
\EndFor
\end{algorithmic}
\end{algorithm}

\vspace{-0.25\baselineskip}\textbf{The State of the Art:} We now discuss the various screening (and aggregation) procedures adopted in different algorithms. The algorithm introduced in~\cite{yin2018byzantine}, termed \emph{robust distributed gradient descent}, uses two screening methods: \emph{coordinate-wise median} and \emph{coordinate-wise trimmed mean}. In coordinate-wise median, the server aggregates the local gradients by taking the median in each dimension of the gradients. Coordinate-wise trimmed mean, on the other hand, involves the server eliminating the smallest and the largest $b$ values in each dimension of the gradients and coordinate-wise averaging the remaining values for aggregation. The GeoMed algorithm~\cite{Chen2017distributed}, in contrast, uses the \emph{geometric median} of local gradients as the screening and aggregation rule. The Krum~\cite{blanchard2017machine} algorithm, on the other hand, finds the local gradient that has the smallest distance to its $M-b-2$ closest gradients and uses this gradient for the update step. There is also a variant of the Krum algorithm, termed \emph{Multi-Krum}~\cite{blanchard2017machine}, which finds $m \in \{1,\dots,M\}$ local gradients using the Krum principle and uses an average of these gradients for update.

The Bulyan algorithm~\cite{mhamdi2018hidden, elmhamdi2019fast} is a two-stage algorithm. In the first stage, it recursively uses vector median methods such as Geometric median and Krum to select $M-2b$ local gradients. In the second stage, it carries out a coordinate-wise operation on the $M-2b$ selected gradients in which $M-4b$ values in each coordinate are retained (and then averaged for aggregation) by eliminating $2b$ values that are farthest from the coordinate-wise median. Zeno/Zeno++~\cite{xie2018zeno, xie2019zeno++} are related algorithms that require an oracle at the server, which can generate an estimate of the true gradient in each iteration, for screening/aggregation purposes. In particular, the screening procedure involves calculating a score for each local gradient based on its difference from the oracle gradient, with the surviving gradients being the ones that are the most similar to the oracle gradient.

The Byzantine-resilient distributed SGD framework of Algorithm~\ref{alg: SGD} has also been investigated in~\cite{xie2018generalized} under a \emph{generalized} Byzantine threat model. The basic assumption underlying this threat model is that the set of nodes under Byzantine attacks can change in each iteration of the algorithm, but the attackers can only inject malicious information in some dimensions of their respective gradients. Similar to \cite{yin2018byzantine,Chen2017distributed}, the work in~\cite{xie2018generalized} puts forth the use of coordinate-wise median, a variant of coordinate-wise trimmed mean, and geometric median for screening purposes, and establishes resilience of the resulting methods under the assumption that less than half of the gradients in each dimension are attacked in each iteration.

We conclude by discussing a few methods that somewhat deviate from the distributed framework of Algorithm~\ref{alg: SGD}. In contrast to \cite{yin2018byzantine,Chen2017distributed,blanchard2017machine,mhamdi2018hidden,elmhamdi2019fast,xie2018zeno,xie2019zeno++,xie2018generalized}, the RSA algorithm of~\cite{li2018rsa} robustifies distributed SGD by making individual nodes store and update local versions of the global optimization variable $\bw$, which are then aggregated at the server in each iteration in a Byzantine-resilient manner. In order to reduce the communications overhead of Byzantine-resilient distributed SGD,~\cite{bernstein2018signsgd} discusses the signSGD algorithm, which only uses the element-wise sign of the gradient---rather than the gradient itself---for the update step. In particular, it is established in~\cite{bernstein2018signsgd} that signSGD with an element-wise majority vote on the signs of the local gradients as a screening/aggregation rule is a Byzantine-resilient learning method. Finally,~\cite{alistarh2018byzantine} proposes and analyzes a variant of distributed SGD that requires only one pass over the entire training data. Unlike other works, however, resilience in this work is accomplished through explicit labeling of nodes as Byzantine, which are then excluded from all future computations at the server.

We next summarize key aspects of some of the Byzantine-resilient distributed learning algorithms in Table~\ref{table:SGD} in terms of the number of nodes $M$, the number of Byzantine nodes $b$, and the number of independent and identically distributed (i.i.d.) samples per node $N$. The convergence rates in the table correspond to the gap between the learned model after $t$ iterations and the minimizer of the statistical risk (cf.~\eqref{eqn: ML problem}) in the limit of large $N$. The parameter $c$ denotes a constant that may change from one algorithm to the other, the $\cO(\cdot)$ scaling hides dependence on problem parameters (including dimension $d$ in the top part of the table), N/A signifies an algorithm lacks a particular result, and --- means guarantees for an algorithm are not directly comparable to other algorithms. The last column in the top part of the table lists conditions on $M$ and $b$ necessary for well-posedness of different algorithms. The bottom part of the table lists per-iteration computational complexity of the screening procedure for algorithms that involve an explicit screening step at the server. While the convergence/learning rates in the table are for strongly convex and smooth loss functions, some of the works require further assumptions and/or also provide results under a relaxed set of assumptions on the loss function and training data. We refer the reader to the referenced works for further details.

\begin{table}
\caption{Summary of some recent results concerning Byzantine-resilient distributed machine learning} \label{table:SGD}
\vspace{-\baselineskip}
\begin{center}
\begin{tabular}{|l|c|c|l|}
\hline
\textbf{Algorithm} & \textbf{Convergence Rate} &\textbf{Statistical Learning Rate} & \textbf{Condition on $(M,b)$}  \\
\hline\hline
Coordinate-wise Median (CM)~\cite{yin2018byzantine} & $\cO\left(c^t\right)$  & $\cO\left(\tfrac{b}{M\sqrt{N}}+\tfrac{1}{\sqrt{MN}}+\tfrac{1}{N}\right)$ &  $M\geq 2b+1$  \\
\hline
Coordinate-wise Trimmed Mean (CTM)~\cite{yin2018byzantine} & $\cO\left(c^t\right)$ & $\cO\left(\tfrac{b}{M\sqrt{N}}+\tfrac{1}{\sqrt{MN}}\right)$ & $M\geq 2b+1$  \\
\hline
GeoMed~\cite{Chen2017distributed} & $\cO\left(c^t\right)$ & $\cO\left(\tfrac{\sqrt{b}}{\sqrt{MN}}\right)$ & $M\geq 2b+1$  \\
\hline
Krum~\cite{blanchard2017machine} & N/A & N/A & $M\geq 2b+3$ \\
\hline
Multi-Krum~\cite{blanchard2017machine} & N/A & N/A & $M\geq 2b+m+2$ \\
\hline
Bulyan~\cite{mhamdi2018hidden} & N/A & N/A & $M\geq 4b+3$ \\
\hline
Zeno/Zeno++~\cite{xie2018zeno, xie2019zeno++} & $\cO\left(c^t\right) + \cO\left(1\right)$ & N/A & $M\geq b + 1$ \\
\hline
RSA~\cite{li2018rsa} & $\cO\left(\tfrac{1}{t}\right) + \cO\left(1\right)$ & N/A & $M\geq b + 1$ \\
\hline
signSGD~\cite{bernstein2018signsgd} & --- & N/A & $M\geq 2b + 1$\\
\hline
\end{tabular}
\end{center}
\vspace{-0.75\baselineskip}
\begin{center}
\begin{tabular}{|c|c|c|c|c|}
\hline
\textbf{Algorithm} & CM, CTM, Zeno/Zeno++ & GeoMed & Krum, Multi-Krum & Bulyan  \\
\hline\hline
\textbf{Screening Complexity} & $\cO(Md)$ & $\cO\left(Md+bd\log^3(\frac{1}{\gamma})\right)^\star$ & $\cO(M^2d)$ & $\cO(M^2d + Md)$ \\
\hline
\end{tabular}
\end{center}
\vspace{-0.25\baselineskip}
\footnotesize{\mbox{${}^\star$ Screening computational complexity for GeoMed is for computing $(1+\gamma)$-approximate geometric median~\cite{Chen2017distributed}.}}
\vspace{-\baselineskip}
\end{table}

\begin{mdframed}[style=MyFrame]
\begin{wrapfigure}{r}{6.5cm}
    \vspace{-\baselineskip}
	\includegraphics[width=6.5cm]{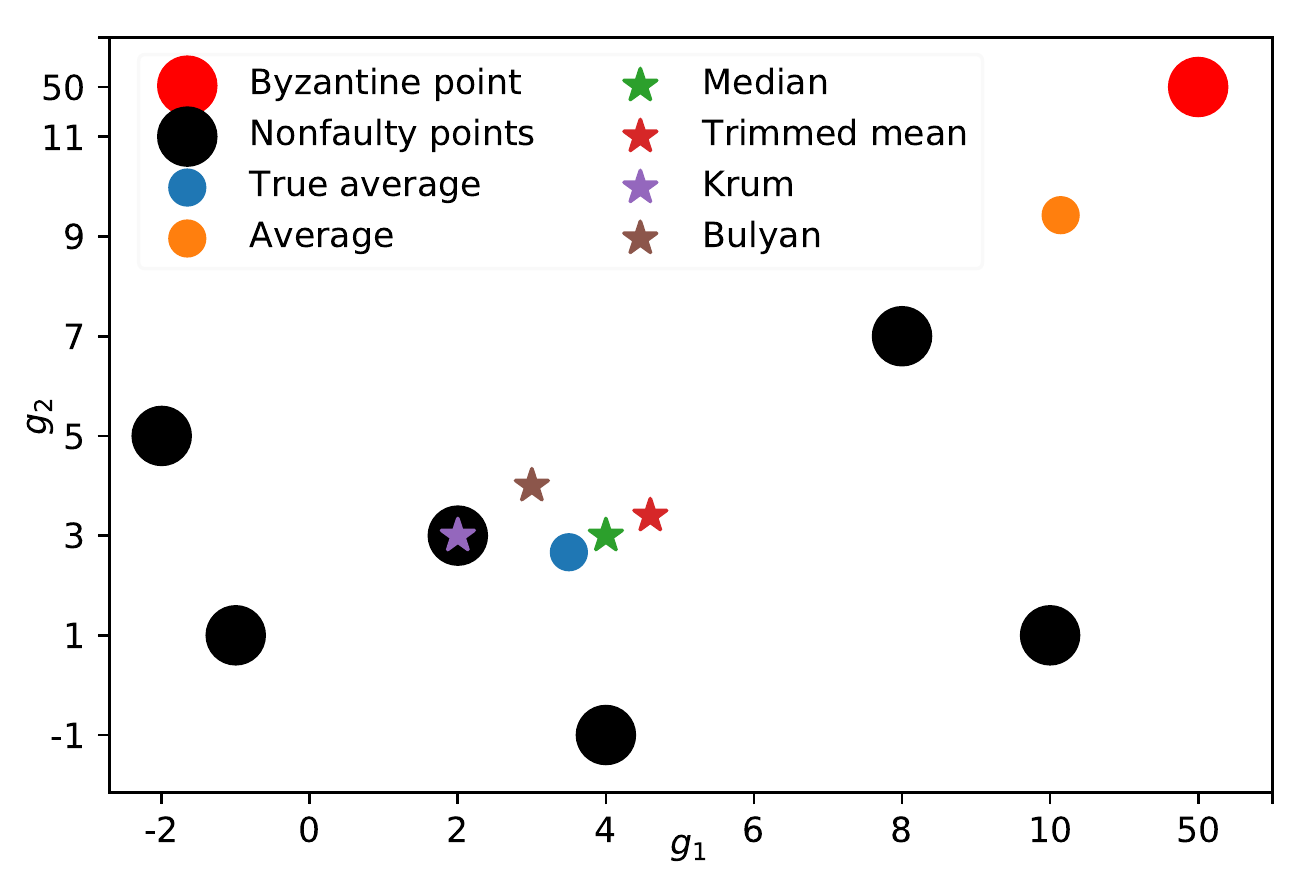}
	\vspace{-1.75\baselineskip}
	\caption{\linespread{1}\selectfont\footnotesize An illustration of the effects of data falsified by Byzantine nodes on screening methods.}
	\label{fig: screening_comparison}
	\vspace{-\baselineskip}
\end{wrapfigure}
\emph{Screening and aggregation in two dimensions:} We illustrate the robustness of different screening/aggregation methods against Byzantine attacks through a simple two-dimensional example. Consider a total of seven two-dimensional points $(g_1, g_2)$, out of which six represent correct data (black discs in Figure~\ref{fig: screening_comparison}) and one of which has been falsified by a Byzantine node (red disc in Figure~\ref{fig: screening_comparison}). It can be seen from Figure~\ref{fig: screening_comparison}, which is displayed with non-uniform $g_1$- and $g_2$-axes to capture the effect of the Byzantine node, that the ordinary average operation is highly susceptible to the falsified data point. On the other hand, screening and aggregation rules such as coordinate-wise median, coordinate-wise trimmed mean, Krum and Bulyan (using Krum in Figure~\ref{fig: screening_comparison}) all produce final results that stay close to the true average of the faultless data points.
\end{mdframed}

\textbf{Numerical Experiments:} It can be seen from Table~\ref{table:SGD} that (coordinate-wise) median, (coordinate-wise) trimmed mean, and GeoMed have some of the best theoretical guarantees for strongly convex and smooth loss functions in terms of convergence and learning rates. We now numerically compare the performance of most of the algorithms listed in Table~\ref{table:SGD} on nonconvex loss functions. Our comparison excludes GeoMed, since it lacks an algorithm for computing the exact geometric median of a set of high-dimensional gradients, as well as signSGD and RSA, since they differ from the rest of the algorithms in terms of their approach to Byzantine resilience.

The nonconvex learning task in the experiments corresponds to classification of the ten classes in CIFAR-10 dataset using a convolutional neural network with three convolution layers followed by two fully connected layers. Each of the three convolution layers are followed by ReLU activation and max pooling, while the output layer uses softmax activation. The distributed setup corresponds to one server and $M=20$ nodes, with the $50,000$ sample training set uniformly at random distributed across the system (i.e., each node has $N = 2,500$ training samples). We run two rounds of experiments for each algorithm, where each round is repeated 10 times---with each trial corresponding to 20,000 algorithmic iterations---and the results on the CIFAR-10 test set are averaged over these 10 trials. In the first round, none of the nodes are taken to be Byzantine nodes. In the second round, four nodes are randomly selected as Byzantine nodes, with each one sending a random vector to the server in each iteration whose elements uniformly take values in the range $(0, 10^{-5})$ for odd-numbered iterations and $(0, 20)$ for even-numbered iterations. We limit ourselves to four Byzantine nodes in order to provide a fairer comparison between different algorithms, since five Byzantine nodes exceeds the theoretical limit of some algorithms (e.g., Bulyan). In both rounds of experiments, the algorithms operate under the assumption of $b = 4$. We conclude our discussion of the experimental setup by noting that the optimal Byzantine attack strategy that adversely affects all distributed learning algorithms in a uniform manner remains an open problem. Nonetheless, the attack strategy being employed in our experiments has been carefully designed in light of the discussions in prior works (see, e.g.,~\cite{xie2019fall}); in particular, it appears to be the uniformly most potent strategy for the distributed learning algorithms under consideration in this article.

\begin{figure}[!t]
    \centering
    \includegraphics[width=\textwidth]{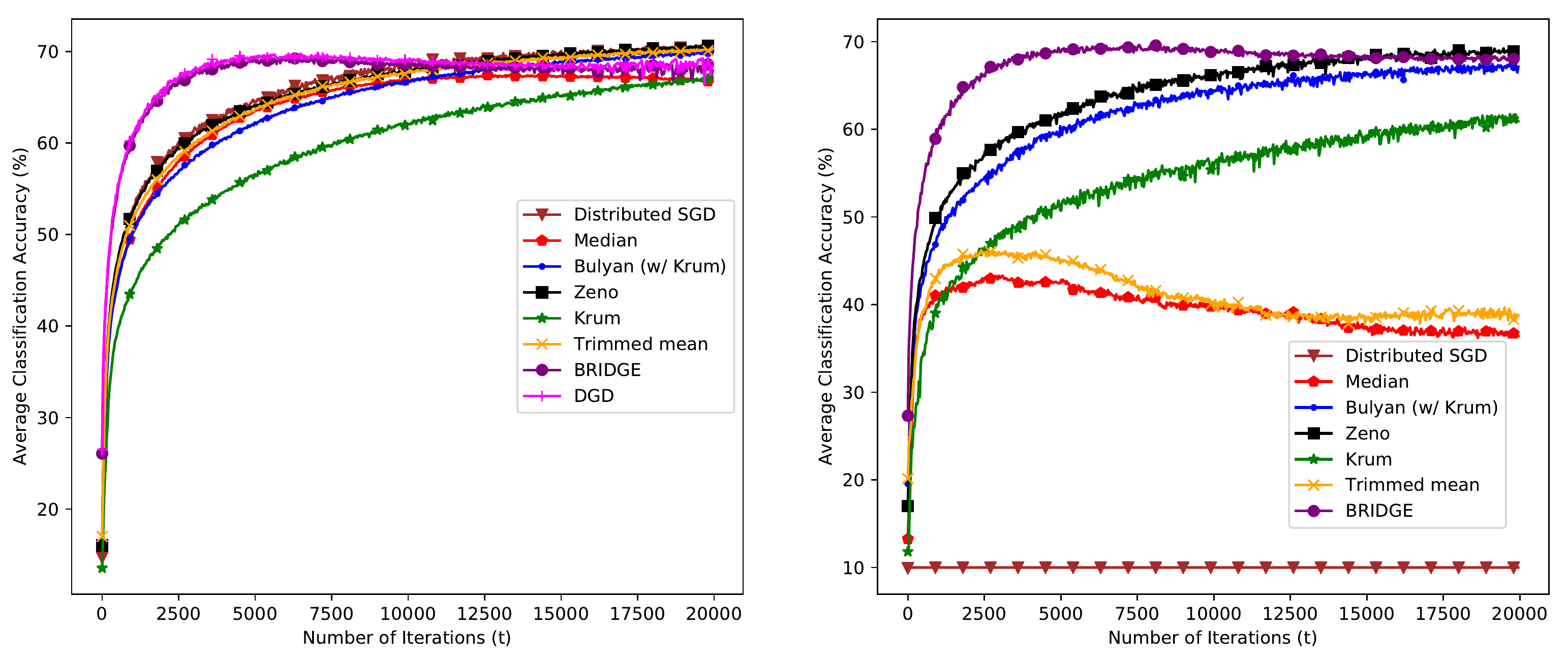}
    \vspace{-1.5\baselineskip}
    \caption{\linespread{1}\selectfont\footnotesize Comparison of different distributed learning methods (with median and trimmed mean being coordinate wise) based on stochastic gradient descent. The left panel corresponds to the faultless setting, while the right panel corresponds to the case of four Byzantine nodes in the system. In both cases, the parameter $b$ is set equal to $4$ in all Byzantine-resilient algorithms. We have also overlayed plots of two decentralized machine learning methods, namely, BRIDGE and DGD, on top of the ones for distributed learning methods; these two methods are discussed later in the article (see ``Decentralized Machine Learning'').}
    \vspace{-0.5\baselineskip}
    \label{fig: distributed_comparison}
\end{figure}

The results of the experiments in terms of average classification accuracy are shown in Figure~\ref{fig: distributed_comparison}, which highlight some trade-offs that should help the practitioners select algorithms that best fit their needs. We first note from Table~\ref{table:SGD} that the computational complexity of screening steps in Krum and Bulyan scales quadratically with the number of nodes $M$, whereas it scales only linearly with $M$ in median, trimmed mean, and Zeno/Zeno++. Next, it can be seen from Figure~\ref{fig: distributed_comparison} that all screening-based methods fall short of distributed SGD's performance in the faultless setting. This is in line with the conventional wisdom that robustness often comes at the expense of correctness, with median and Krum paying the highest price in terms of correctness, while Bulyan, trimmed mean, and Zeno paying the least (and somewhat insignificant) price. In the presence of Byzantine nodes, on the other hand, the vanilla distributed SGD completely falls apart. In contrast, while varying degrees of reduction in the performance of screening-based methods are observed under Byzantine attacks, none of them breaks down to the level of distributed SGD. Nonetheless, median, trimmed mean, and Krum (in this particular order) are affected the most in terms of performance by the aggressive Byzantine attack strategy employed in our experiments. Bulyan and Zeno, on the other hand, seem to be the most stable under Byzantine attacks. In addition, both these methods offer a competitive tradeoff between correctness (in a faultless setting) and robustness (under attack). However, it is worth pointing out that Zeno's resilience comes at the expense of an oracle that can provide it with some knowledge of the true gradient. Similarly, since Bulyan screens four times the number of Byzantine nodes in each iteration, its resilience comes at the expense of a limited number of Byzantine nodes that it can handle (cf.~Table~\ref{table:SGD}).

\section{Adversary-resilient Decentralized Processing of Data}
Distributed systems in general, and distributed master--worker architectures in particular, have become the workhorse framework for non-centralized processing of data. The reasons for this range from relative ease of implementation and subsequent scaling up or scaling down of the system through addition or removal of nodes to relative simplicity of synchronization and communication protocols. Despite these and other related advantages, decentralized setups for inference and learning are increasingly being investigated in lieu of distributed setups for a multitude of reasons. Unlike the distributed setup, a decentralized system lacks a central server that is connected to every node (see ``Decentralized Setup'' in Figure~\ref{fig:distributed.vs.decentralized}). Instead, all nodes in a decentralized system maintain local copies of the decision variable/machine learning model and reach (approximate) consensus on a common solution through periodic exchange of messages over a network with a subset of other nodes, termed \emph{neighbors}. Because of this reason, and unlike distributed systems, there is no single point of failure in decentralized systems. In the same vein, whereas the star communications topology in distributed master--worker architectures can create a communications bottleneck at the central server (see ``Distributed Setups'' in Figure~\ref{fig:distributed.vs.decentralized}), the network/communications topologies in decentralized systems can be carefully designed to avoid such bottlenecks. One of the biggest reasons for the study of Byzantine-resilient inference and learning in decentralized systems, however, is the emergence of inherently decentralized applications such as networks of self-driving cars and robot swarms. Both centralized and distributed setups typically cannot be engineered in a cost-effective manner in such applications, which are generally studied under the moniker of \emph{multiagent systems}. This necessitates the use of decentralized frameworks, often with ad-hoc network topologies, for statistical inference and machine learning.

Since no single node in decentralized setups gets access to the entire set of local variables, ensuring robustness against Byzantine failures is generally harder in decentralized systems. In particular, unlike simple characterizations of the feasible and/or necessary relationships between the number of nodes $M$ and the number of Byzantine nodes $b$ in Table~\ref{table: distributed_detection} and Table~\ref{table:SGD}, the necessary and/or sufficient conditions for Byzantine-resilient decentralized algorithms are stated in terms of topology of the underlying network graph. In addition, while practical benefits of asynchronous update rules for and impact of time-varying network topologies on decentralized processing of data have long been investigated by researchers, the literature on Byzantine-resilient decentralized processing is relatively sparse in this regard. Because of this reason, and due to space constraints, our discussion in this article is mostly limited to synchronous algorithms on decentralized networks whose topology does not change with time (i.e., \emph{static graphs}). We begin by engaging in discussion of an algorithmic process, aptly termed \emph{consensus}, that often forms the basis of algorithms for decentralized inference and machine learning.

\subsection{Decentralized Consensus}
Many decentralized inference and learning algorithms require a subprocess that ensures (approximate) agreement, i.e., consensus, among all nodes. In the context of Byzantine-resilient decentralized processing, therefore, it is instructive to understand the fundamentals of Byzantine-resilient consensus. Our discussion in this regard will be limited to linear strategies for consensus, which form the basis for Byzantine-resilient decision consensus, averaging consensus, convex consensus, etc.~\cite{Leblanc2013resilient, Vaidya2012matrix}.

Suppose each node $j$ in the decentralized system has a local variable $\bw_j^0$ and it is desired to reach consensus on the average of these variables at each node through in-network message passing, i.e., $\bw_j \approx \tfrac{1}{M}\sum\limits_{i=1}^M \bw_i^0$ for every node $j$. Consensus algorithms usually proceed by having each node iteratively take a weighted average of their neighbors' variables. Mathematically, therefore, a consensus algorithm is associated with an $M \times M$ \emph{weight matrix} whose $(j,i)$-th entry corresponds to the weight assigned by node $j$ to the variable it receives from node $i$. In the absence of Byzantine nodes, a well-known result says that a doubly stochastic weight matrix whose smallest nonzero entry is lower bounded guarantees convergence of all $\bw_j$'s to the average $\tfrac{1}{M}\sum\limits_{i=1}^M \bw_i^0$. Suppose, however, that node $k$ in the system is Byzantine and it transmits $\bw_k = \bw'$ to its neighbors in each iteration. This simple ``lazy'' Byzantine strategy is enough for all nodes to converge to $\bw'$. And if there were another Byzantine node $k'$ in the system that always transmitted the same $\bw_{k'} \neq \bw'$, nodes in the system will not reach consensus at all.

The main idea behind ensuring robustness of (averaging) consensus to Byzantine nodes is to screen potential outliers at each node before the weighting step in each iteration. Since averaging of finite-dimensional vectors is equivalent to averaging of their individual respective (scalar) coordinates, much of the discussion in Byzantine-resilient consensus has focused on scalar-valued problems. Some variant of the trimmed mean, in which a node removes the largest $b$ and the smallest $b$ values received from its neighbors (including itself), is in particular a common screening method used in scalar consensus. Clearly, each node has to have enough neighbors (e.g., more than $2b$ for trimmed mean) for screening-based consensus to be well defined. However, since the network must not become disconnected after screening, the exact constraint on network topology for different algorithms tends to be more involved; see \cite{Leblanc2013resilient, Vaidya2012matrix} for further discussion of different screening methods and the corresponding topology constraints. Note that decentralized inference and learning algorithms utilizing similar screening ideas for Byzantine resilience tend to have similar topology constraints.

In terms of the performance of Byzantine-resilient consensus, even the best screening/aggregation method cannot guarantee removal of all falsified data. However, equipped with a reasonable strategy, a Byzantine-resilient scalar consensus algorithm can guarantee two things in each iteration: ($i$) the retained values are between the smallest and the largest scalars at all nonfaulty nodes; and ($ii$) the difference between the largest and the smallest nonfaulty values decreases. Together, these two conditions can ensure that the nonfaulty nodes converge to a common value. However, since a node can also retain falsified data and/or eliminate nonfaulty data, consensus to the true average is no longer possible. Instead, guarantees can only be given to an approximate average such as a convex combination of initial values. This in particular is what makes Byzantine-resilient decentralized inference and learning so challenging.

We conclude by noting that asynchronous variants of decentralized consensus in the presence of Byzantine attacks have been investigated in~\cite{haseltalab2015approximate,silvestre2017stochastic}. The asynchronicity in~\cite{silvestre2017stochastic} comes through the use of \emph{randomized gossip} algorithms, but the focus in that work is primarily on detection of Byzantine nodes. In contrast, \cite{haseltalab2015approximate} extends the synchronous (screening) framework and analysis of~\cite{Leblanc2013resilient} to asynchronous settings (and time-varying network topologies). A noteworthy implication of~\cite{haseltalab2015approximate} is that asynchronicity does not impose additional topology constraints required for resilience of decentralized consensus algorithms. This is important since synchronous algorithms often incur additional latency; indeed, a synchronous algorithm can only be as fast as the slowest connection/node in the system.

\subsection{Decentralized Statistical Inference}
Similar to the case of distributed inference, we limit ourselves in here to decentralized detection and estimation. In both problems, decentralized consensus makes an integral part of the developed algorithms.

\textbf{Decentralized Detection:} Similar to distributed detection, a typical setup in decentralized detection involves nodes in the system making observations under one of two (or more) hypotheses. Unlike distributed detection, however, nodes must collaborate among themselves to reach consensus in favor of one of the hypotheses. Vulnerability of the vanilla consensus framework to Byzantine nodes, therefore, also makes decentralized detection highly susceptible to Byzantine attacks.

While several application scenarios call for Byzantine-resilient decentralized detection, it has received less attention compared to its distributed counterpart. Some of the most relevant works in this regard include~\cite{zhu.guan.globecom2012, kalikhura.goldhahn2017, kalikhura.varshney2017}, all of which focus on scalar-valued problems. In the spirit of Byzantine-resilient scalar consensus, an adaptive threshold-based screening method is utilized in~\cite{zhu.guan.globecom2012} to mitigate the impact of Byzantine nodes on the final decision. Similarly, a robust variant of distributed \emph{alternating direction method of multipliers} (ADMM) is proposed in~\cite{kalikhura.goldhahn2017} that uses trimmed mean to eliminate $2b$ scalars at each node in every iteration. But other than the trivial topology constraint imposed by the trimmed mean (i.e., in-degree of nodes being greater than $2b$), this work lacks guarantees. In contrast, \cite{kalikhura.varshney2017} proposes and analyzes a robust detection scheme that involves identification of Byzantine nodes and use of a weighted average consensus algorithm. However, this work focuses on a particular variant of Byzantine attacks and does not characterize topology constraints as a function of the number of Byzantine nodes.

\textbf{Decentralized Estimation:} Decentralized estimation, in which nodes use local observations and messages from neighbors to reach consensus on the estimate of an unknown parameter $\bw \in \R^d$, is often also studied under the linear model. Similar to decentralized detection, Byzantine-resilient decentralized estimation is a relatively recent topic of interest. Among relevant works, \cite{LeBlanc.hassan2014} focuses on a specialized estimation problem in which node $j$ need only estimate the $j$-th entry of $\bw$ and there are three types of nodes in the system: \emph{reliable} nodes, which are special nodes having perfect knowledge of the entry of $\bw$ associated with them, ordinary (normal) nodes, and Byzantine nodes. Under this setup and the assumption of a time-varying directed graph, \cite{LeBlanc.hassan2014} proposes and analyzes a trimmed mean-type (scalar-valued) screening procedure for Byzantine resilience. The works~\cite{kar.moura2018trans, kar.moura2018} study vector-valued dynamic estimation under the noiseless observation model of $\by_j[n] = \bw, \ n=1,\dots,$ for nonfaulty nodes. While the threat model in~\cite{kar.moura2018trans, kar.moura2018} is a specialized variant of the Byzantine model, it allows the indices of the nodes under attack to change from one time instance to the next one. Both~\cite{kar.moura2018trans, kar.moura2018} use similar consensus+innovations algorithms coupled with time-varying gains to achieve resilience. The main difference in these works is that~\cite{kar.moura2018} can achieve linear convergence, as opposed to sublinear convergence for~\cite{kar.moura2018trans}, but it can only tolerate less than $30\%$ of the nodes being under attack, in contrast to $50\%$ for~\cite{kar.moura2018trans}. Finally, \cite{chen.kar.moura2018} accomplishes Byzantine-resilient decentralized dynamic estimation under the general linear model $\by_j[n] = \bH_j \bw + \bEta_j[n]$ by focusing on explicit detection of adversaries. We conclude by noting that the review article~\cite{chen.moura2018article} provides more detailed overview of~\cite{LeBlanc.hassan2014, kar.moura2018trans, chen.kar.moura2018}.

\subsection{Decentralized Machine Learning}
Decentralized machine learning algorithms, which can be considered a combination of consensus and distributed learning frameworks, approximately solve~\eqref{eqn: ML problem} by minimizing a \emph{global} loss function on the non-collocated data in a decentralized manner while reaching an agreement among all nodes, i.e.,
\begin{align}\label{eqn: decentralized learning}
    \min\limits_{\{\bw_1,\dots,\bw_M\}} \frac{1}{M}\sum\limits_{j=1}^M f(\bw_j, \bZ_j) \quad \text{subject to} \quad \bw_i = \bw_j \ \forall i,j,
\end{align}
where $\bw_j \in \R^d$ denotes the model learned at node $j$ in the system. We refer to \eqref{eqn: decentralized learning} as the \emph{decentralized} ERM problem, which approximately solves the statistical risk minimization problem \eqref{eqn: ML problem}.

We focus here on synchronous gradient descent-based methods for solving \eqref{eqn: decentralized learning}. The classic \emph{decentralized gradient descent} (DGD) method~\cite{Nedic2009distributed}, for instance, involves each node $j$ exchanging its current local iterate in every iteration $t$ with all nodes in its neighborhood $\cN_j$ and then updating the local iterate using a consensus-type weighted averaging step and a local gradient descent step, i.e.,
\begin{align}
    \label{eqn:DGD.iteration}
    \bw_j^{t+1} = \underbrace{\alpha_{jj} \bw_j^t + \sum\limits_{i\in \cN_j}\alpha_{ji} \bw_i^t}_{\text{consensus}} \underbrace{- \rho(t) \bg_j(\bw_j^t)}_{\text{local gradient descent}},
\end{align}
where $\{\alpha_{ji}\}$ is the collection of averaging weights, $\rho(t)$ denotes the step size, and $\bg_j(\bw) := \nabla_\bw f(\bw, \bZ_j)$ denotes the local gradient. Similar to the case of ``Decentralized Consensus,'' however, a simple lazy Byzantine strategy at one or more nodes will lead to breakdown of DGD-based learning methods.

\textbf{Scalar-valued Problems:} Robustification of decentralized ERM-based learning in the presence of Byzantine attacks requires resilient variants of DGD. The works in~\cite{Su2015fault, sundaram2018distributed} focus on this for scalar-valued (i.e., $d=1$) decentralized optimization. Similar to distributed learning and decentralized consensus, resilience in these methods is also achieved through the use of a screening-based aggregation procedure within the consensus step in \eqref{eqn:DGD.iteration}. The works~\cite{Su2015fault, sundaram2018distributed} in particular focus on trimmed-mean screening: after node $j$ receives $w_i \in \R$ from its neighbors, it removes the largest $b$ and the smallest $b$ $w_i$'s and takes average of the remaining scalars for consensus. This leads to the following update rule:
\begin{align}\label{eqn: scalar update}
	w_j^{t+1} =\frac{1}{|\cN_{j^*}^t|} \sum\limits_{i\in \cN_{j^*}^t}w_i^t - \rho(t)g_j(w_j^t),
\end{align}
where $\cN_{j^*}^t$ is the set of nodes (possibly including node $j$ itself) that survive the screening at node $j$. This update can be shown to be Byzantine resilient in the sense that the $w_j^t$'s converge to the minimizer of \emph{some} convex combination of local losses $f(w,\bZ_j)$. We also have a result from~\cite{Su2015fault} that the minimum of the decentralized ERM in \eqref{eqn: decentralized learning}, even when restricted to the set of nonfaulty nodes, cannot be achieved in the presence of Byzantine nodes. Despite this negative result, it can be shown that any convex combination of local losses will converge in probability to the global statistical risk in the case of i.i.d. training samples. This can then be leveraged to establish the robustness of \eqref{eqn: scalar update} for scalar-valued decentralized learning~\cite{Yang2017byrdie}.

\textbf{Vector-valued Problems:} The algorithms in~\cite{Su2015fault, sundaram2018distributed} cannot be directly utilized in vector-valued problems (i.e., $\bw \in \R^d$ for $d > 1$). On the one hand, unless a problem decouples over different coordinates of the optimization variable $\bw$, minimizing the objective function along one coordinate \emph{independent} of the other coordinates does not yield the right solution. On the other hand, since the trimmed-mean procedure of~\cite{Su2015fault, sundaram2018distributed} requires sorting of values received from one's neighbors, it cannot be directly applied to members of an \emph{unordered space} like $\R^d$. This limitation of~\cite{Su2015fault, sundaram2018distributed} is overcome in~\cite{Yang2017byrdie}, which proposes an algorithm termed \emph{Byzantine-resilient decentralized coordinate descent} (ByRDiE) for vector-valued decentralized learning in the presence of Byzantine nodes. ByRDiE, fundamentally being a coordinate descent method, cyclically updates one coordinate at a time in a decentralized manner. And since each subproblem in coordinate descent becomes a scalar-valued problem, ByRDiE uses trimmed-mean screening in each inner (coordinate-wise) iteration for Byzantine resilience. The final update rule in each coordinate-wise iteration of ByRDiE takes a form similar to \eqref{eqn: scalar update}, except that the gradient term also depends on other coordinates of $\bw$. For strictly convex and smooth loss functions, \cite{Yang2017byrdie} guarantees (algorithmic and statistical) convergence of the iterates of ByRDiE to the statistical minimizer, with the algorithmic convergence rate being sublinear and the statistical learning rate being $\cO\left(a/\sqrt{MN}\right)$; here, $N$ again denotes the number of i.i.d. training samples per node, while the parameter $a$ differs from one problem setup (and Byzantine attack model) to another and satisfies $1 \leq a \leq \sqrt{M}$. Despite these guarantees, which establish that ByRDiE can result in robust and (sample-wise) fast statistical learning in decentralized setups that exceeds the local learning rate of $\cO\left(1/\sqrt{N}\right)$~\cite{Yang2017byrdie}, the one-coordinate-at-a-time update of ByRDiE can be inefficient for high-dimensional (i.e., $d \gg 1$) problems. Indeed, since the coordinate-wise gradient update step depends on the updates of other coordinates, the iterates in ByRDiE cannot be updated in a coordinate-wise parallel fashion, leading to high network coordination and local computation overheads in decentralized learning.

To curtail the high overheads of ByRDiE,~\cite{yang2019bridge} presents another algorithm---termed \emph{Byzantine-resilient decentralized gradient descent} (BRIDGE)---that is based on gradient descent and coordinate-wise trimmed mean. Similar to DGD, each node $j$ in BRIDGE also exchanges its \emph{entire} current iterate $\bw_j^t$ in every iteration $t$ with all nodes in its neighborhood $\cN_j$. The update step in BRIDGE, however, involves coordinate-wise screening/aggregation of $\bw_j^t$'s using trimmed mean with parameter $2b$, which is followed by a local gradient descent step. Mathematically, the (parallel) update of the $k$-th coordinate is given by
\begin{align}
	\forall k \in \{1,\dots,d\} \ \textsf{(in parallel)}, \ w_j^{t+1}(k) =\frac{1}{|\cN_{j^*}^{t,k}|} \sum\limits_{i\in \cN_{j^*}^{t,k}} w_i^t(k) - \rho(t) g_j^k(\bw_j^t),
\end{align}
where $\cN_{j^*}^{t,k}$ is the set of nodes whose $k$-th coordinates survive the screening at node $j$ and $g_j^k(\bw)$ denotes the $k$-th coordinate of the local gradient $\bg_j(\bw)$. In the case of strongly convex and smooth loss functions, theoretical guarantees for BRIDGE match those for ByRDiE.

\textbf{Topology Constraints:} We noted earlier (see ``Decentralized Consensus'') that Byzantine resilience in decentralized setups depends on network topology. We now describe two related topology constraints for trimmed mean-based decentralized learning. The constraint in \cite{Su2015fault} requires that, after removing all $b$ Byzantine nodes and \emph{any} combination of remaining $b$ (incoming) edges from each node, there is always a group of nodes---termed \emph{source component}---of cardinality at least $(b+1)$ that has a directed path to every other node. In words, this constraint means every nonfaulty node, even after trimmed-mean screening, can always receive information---directly or indirectly---from the source component. The topology constraints for ByRDiE and BRIDGE are also based on this condition. The constraint in \cite{sundaram2018distributed}, on the other hand, is based on the idea that any two arbitrary partitions of the network must result in one of the partitions having at least one node with $(2b+1)$ neighbors outside the partition. This, in turn, guarantees that $b$ Byzantine nodes cannot isolate any subset of nonfaulty nodes during trimmed-mean screening.

\textbf{Beyond Trimmed-mean Screening:} Unlike distributed learning, where several screening methods and aggregation rules have been proposed and analyzed for Byzantine resilience (Table~\ref{table:SGD}), utilization of screening methods in decentralized learning has been limited to (coordinate-wise) trimmed mean. The main reason for this is the need for iterate consensus in decentralized learning, which makes analysis of other screening/aggregation methods challenging. (It is also worth noting that our focus here is on iterate screening, as opposed to gradient screening in distributed learning.) In practice, it is possible to merge the ideas behind BRIDGE and screening methods such as coordinate-wise median, Krum, and Bulyan to develop other Byzantine-resilient decentralized learning methods. This is indeed the approach we have taken in the numerical experiments discussed in the following. Consensus, convergence analysis, and statistical learning rates of such methods, however, remain an open problem.

\textbf{Numerical Experiments:} We first compare the performance of DGD, ByRDiE, BRIDGE and three variants of BRIDGE that come about from incorporation of the screening principles of (coordinate-wise) median, Krum, and Bulyan (with Krum) within the BRIDGE framework. The computational inefficiency of ByRDiE for high-dimensional datasets and machine learning models necessitates an experimental setup, both in terms of the dataset and the learning task, that differs from the one utilized as part of our discussion on Byzantine-resilient distributed machine learning. Specifically, note that each (colored) image in the CIFAR-10 dataset is $3,072$-dimensional and the corresponding convolutional neural network used earlier in our discussion for this dataset gives rise to a model with $d = 122,410$. In contrast, we resort to a computationally tractable experimental setup in here that corresponds to multiclass classification of MNSIT dataset ($784$-dimensional data samples) using a linear one-layer neural network (i.e., $d = 7,840$). In addition to being computationally manageable for ByRDiE, this setup also results in a loss function that satisfies the theoretical conditions for convergence of ByRDiE and BRIDGE.

\begin{figure}
    \centering
    \includegraphics[width=\textwidth]{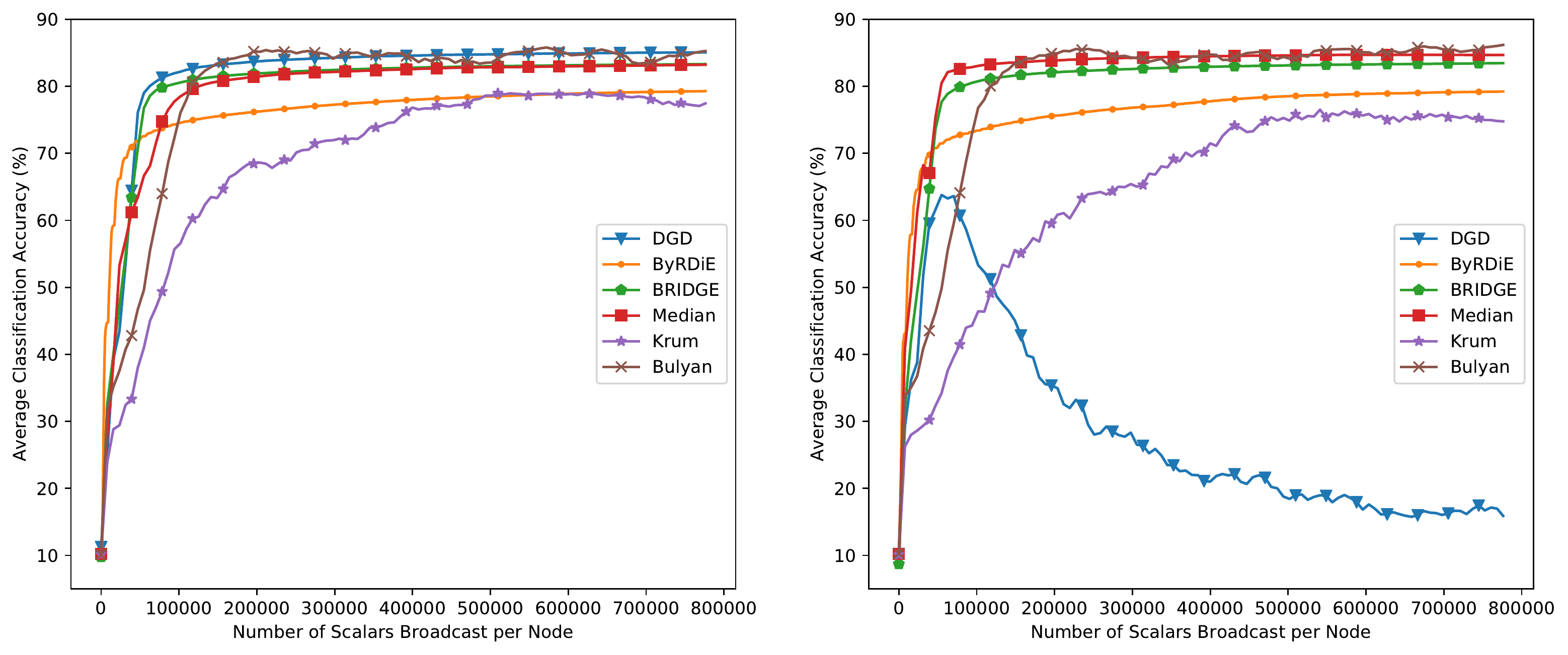}
    \vspace{-1.5\baselineskip}
    \caption{\linespread{1}\selectfont\footnotesize Performance comparison of different decentralized learning methods in both faultless (left panel) and Byzantine settings (right panel). Byzantine-resilient algorithms in both settings operate under the assumption of $b=2$. The algorithms entitled \emph{Median}, \emph{Krum}, and \emph{Bulyan} are effectively BRIDGE combined with the screening procedures advocated in distributed learning.}
    \vspace{-0.5\baselineskip}
    \label{fig: decentralized_comparison}
\end{figure}

The decentralized system in our experimental setup involves a total of $M=20$ nodes in the network, with a communications link (edge) between two nodes decided by a random coin flip. Once a random topology is generated, we ensure each node has at least $4b+1$ nodes in its neighborhood (a condition imposed due to Bulyan screening). The training data at each node corresponds to $N=2,000$ samples randomly selected from the MNIST dataset. The performance of each method is reported in terms of classification accuracy, averaged over $(M-b)$ nodes and a total of 10 independent Monte Carlo trials, as a function of the number of scalars broadcast per node. The final results, shown in Figure~\ref{fig: decentralized_comparison}, correspond to two sets of experiments: ($i$) the faultless setting in which none of the nodes actually behaves maliciously; and ($ii$) the setting in which two of the $20$ nodes are Byzantine, with each Byzantine node broadcasting every coordinate of the iterate as a uniform random variable between $-1$ and $0$. Note that this Byzantine attack strategy is by no means the most potent in all decentralized settings. However, similar to the attack strategy utilized in our earlier discussion on distributed learning, this particular strategy has been selected after careful evaluation of the impact of different strategies proposed in works such as \cite{xie2019fall,fang2019local,baruch2019little} on our particular experimental setup. Finally, with the exception of DGD, all methods are initialized with parameter $b=2$ in both faultless and faulty scenarios.

It can be seen from Figure~\ref{fig: decentralized_comparison} that, other than ByRDiE and Krum-based screening, all methods perform almost as well as DGD in the faultless case. In the presence of Byzantine nodes, however, DGD completely falls apart, whereas the performances of all screening methods remain comparable to the faultless setting. It is also worth comparing these results to those for Byzantine-resilient distributed learning (see Figure~\ref{fig: distributed_comparison}). While (coordinate-wise) median and trimmed mean appear to be the worst performers in Figure~\ref{fig: distributed_comparison}, Krum-based screening is the least effective in Figure~\ref{fig: decentralized_comparison}. In both cases, however, Bulyan is quite effective, except that it has stringent topology requirements.

We conclude by explicitly comparing the performance of DGD and BRIDGE in decentralized settings to that of (Byzantine-resilient) learning methods in distributed settings. Since DGD and BRIDGE are scalable to high-dimensional learning tasks, our experimental setup, dataset, data distribution, learning task, and Byzantine attack strategy for this comparison are identical to the ones described earlier for Byzantine-resilient distributed learning methods. In order to ensure the decentralized setup in the case of BRIDGE satisfies the topology constraints corresponding to four Byzantine nodes in the system, we use $0.7$ as the probability of random connectivity between any two nodes. The final results for DGD and BRIDGE, which are overlayed on top of the ones for distributed learning methods in Figure~\ref{fig: distributed_comparison}, show that BRIDGE offers competitive performance in both faultless and faulty systems. In fact, BRIDGE (and DGD) have faster convergence rates than the distributed learning methods for this particular nonconvex problem, decentralized setup, and Byzantine strategy. While this could be partly attributable to higher network connectivity of $0.7$ in the decentralized setting, a careful head-to-head comparison and understanding of distributed and decentralized learning methods in both faultless and faulty settings is an open problem.

\section{Some Open Research Problems}
Despite recent advances, Byzantine-resilient inference and learning remains an active area of research with several open problems. Much of the focus in distributed inference has been on the somewhat restrictive model in which Byzantine nodes do not collude. Collaborative Byzantine attacks, on the other hand, can be much more potent than independent ones. A fundamental understanding of mechanisms for safeguarding against such attacks remains a relatively open problem in distributed inference. Byzantine-resilient distributed estimation under nonlinear models is another problem that has been relatively unexplored. In the case of Byzantine-resilient distributed learning, existing works have only scratched the surface. Convergence and/or learning rates of many of the proposed methods remain unknown (Table~\ref{table:SGD}). In addition, while SGD is a workhorse of machine learning, approaches such as accelerated first-order methods (e.g., accelerated gradient descent), first-order dual methods (e.g., ADMM), and second-order methods (e.g., Newton's method) do play important roles in machine learning. However, resilience of distributed variants of such methods to Byzantine attacks has not been investigated in the literature.

The lack of a central server, the need for consensus, and an ad-hoc topology make it even more challenging to develop and analyze Byzantine-resilient methods for decentralized inference and learning. Much of the work in this regard is based on screening methods such as trimmed mean and median that originated in the literature on Byzantine-resilient scalar-valued consensus. This has left open the question of how other screening methods, such as the ones explored within distributed learning, might handle Byzantine attacks---both in theory and in practice---in various decentralized problems. Unlike distributed learning, any such efforts will also have to characterize the interplay between network topology and effectiveness of the screening procedure. The fundamental tradeoffs between the robustness and the (faultless) performance of Byzantine-resilient methods also remain largely unknown for decentralized setups. Finally, existing works on decentralized learning only guarantee sublinear convergence for strictly/strongly convex and smooth functions. Whether this can be improved by taking advantage of faster distributed optimization frameworks or different screening methods also remains an open question.

\section{Conclusions}
In this article, we have presented an overview of latest advances in Byzantine-resilient inference and learning. In the distributed master--worker setting, which is characterized by the presence of a central server that computes the final solution, we discussed recent results concerning resilience of distributed detection, estimation, and machine learning against Byzantine attacks. Within distributed machine learning, we focused on Byzantine-resilient variants of distributed stochastic gradient descent, whose performances were compared using numerical experiments. In the decentralized setting, which typically requires consensus due to lack of a central server, we first discussed the principles behind Byzantine-resilient consensus. This was followed by a discussion of latest results on decentralized detection, estimation, and learning in the presence of Byzantine nodes. We also compared and contrasted different Byzantine-resilient decentralized learning methods using numerical experiments, and discussed similarities and differences between them and distributed learning methods. Byzantine-resilient inference and learning has a number of research challenges that remain unaddressed, some of which are also briefly discussed in the article.

\section*{Acknowledgements}
The authors gratefully acknowledge the support of the NSF (CCF-1453073, CCF-1907658), the ARO (W911NF-17-1-0546), and the DARPA Lagrange Program (ONR/SPAWAR contract N660011824020).

\section*{Short Biographies}
\noindent\textbf{\emph{Zhixiong Yang}} (zhixing.yang@rutgers.edu) received BS degree from Beijing Jiaotong University, China in 2011, MS degree from Northeastern University, MA in 2013, and PhD in electrical engineering from Rutgers University--New Brunswick, NJ in 2020. He is currently a system engineer at Blue Danube Systems. His research interests are distributed processing, machine learning, and massive MIMO systems.

\noindent\textbf{\emph{Arpita Gang}} (arpita.gang@rutgers.edu) received BTech degree from the National Institute Of Technology, Silchar, India, and MTech degree from IIIT-Delhi, India. Since 2017, she has been pursuing PhD in electrical engineering from Rutgers University--New Brunswick, NJ. Her research interests are distributed optimization and signal processing. She is a graduate student member of the IEEE.

\noindent\textbf{\emph{Waheed U. Bajwa}} (waheed.bajwa@rutgers.edu) has been with Rutgers University--New Brunswick, NJ since 2011, where he is currently an associate professor in the Departments of Electrical \& Computer Engineering and Statistics. His research interests include statistical signal processing, high-dimensional statistics, and machine learning. He is a senior member of the IEEE.

\bibliographystyle{IEEEbib}

\end{document}